\documentclass[nohyperref]{article}

\usepackage{microtype}
\usepackage{graphicx}
\usepackage{subfigure}
\usepackage{booktabs} 

\usepackage{hyperref}

\newcommand{\minus}{\scalebox{0.7}[1.0]{$-$}}

\usepackage[accepted]{icml2022_arxiv}

\usepackage{amsmath}
\usepackage{amssymb}
\usepackage{mathtools}
\usepackage{amsthm}
\usepackage{stackengine}
\usepackage{setspace}

\usepackage[capitalize,noabbrev]{cleveref}

\theoremstyle{plain}

\theoremstyle{definition}

\theoremstyle{remark}

\usepackage[textsize=tiny]{todonotes}

\icmltitlerunning{The unreasonable effectiveness of few-shot learning for machine translation}

\begin{document}

\twocolumn[
\icmltitle{The unreasonable effectiveness of few-shot learning for machine translation}



\icmlsetsymbol{equal}{*}

\begin{icmlauthorlist}
\icmlauthor{Xavier Garcia}{res}
\icmlauthor{Yamini Bansal}{res}
\icmlauthor{Colin Cherry}{tr}
\icmlauthor{George Foster}{tr}
\icmlauthor{Maxim Krikun}{tr}
\icmlauthor{Fangxiaoyu Feng}{tr}
\icmlauthor{Melvin Johnson}{res}
\icmlauthor{Orhan Firat}{res}

\end{icmlauthorlist}

\icmlaffiliation{res}{Google Research - Brain Team}
\icmlaffiliation{tr}{Google Translate}

\icmlcorrespondingauthor{Xavier Garcia}{xgarcia@google.com}
\icmlcorrespondingauthor{Orhan Firat}{orhanf@google.com}

\icmlkeywords{Machine Learning}

\vskip 0.3in
]



\printAffiliationsAndNotice{}  

\begin{abstract}
We demonstrate the potential of few-shot translation systems, trained with unpaired language data, for both high and low-resource language pairs. We show that with only 5 examples of high-quality translation data shown at inference, a transformer decoder-only model trained solely with self-supervised learning, is able to match specialized supervised state-of-the-art models as well as more general commercial translation systems. In particular, we outperform the best performing system on the WMT'21 English\minus{}Chinese news translation task by only using five examples of English\minus{}Chinese parallel data at inference. Moreover, our approach in building these models does not necessitate joint multilingual training or back-translation, is conceptually simple and shows the potential to extend to the multilingual setting. Furthermore, the resulting models are two orders of magnitude smaller than state-of-the-art language models. We then analyze the factors which impact the performance of few-shot translation systems, and highlight that the quality of the few-shot demonstrations heavily determines the quality of the translations generated by our models.  Finally, we show that the few-shot paradigm also provides a way to control certain attributes of the translation --- we show that we are able to control for regional varieties and formality using only a five examples at inference, paving the way towards controllable machine translation systems.  
\end{abstract}

\section{Introduction}

Current state-of-the-art machine translation systems are typically built by leveraging vast amounts of parallel data mined from the web. While this is practical for high-resource language pairs, it is unfeasible to obtain corpora of such sizes for the majority of languages in the world. Moreover, the reliance of mined parallel data has many potential downsides, such as allowing for poisoning attacks \cite{xu2021targeted,wang2021putting}, memorization of low-quality examples \cite{raunak2022finding}, and biases towards generating text in language registers over-represented in the parallel data, such as language varieties \cite{lakew2018neural,riley2022frmt} or formality \cite{rippeth-etal-2022-controlling}.

As an alternative, researchers began exploring the task of \emph{unsupervised translation} \cite{ravi2011deciphering} i.e. building translation models without any parallel data at all. Unsupervised translation systems have demonstrated promising performance in recent years, able to match strong supervised baselines on academic benchmarks \cite{song2019mass,garcia2020harnessing,han2021unsupervised} by relying on a collection of tricks and techniques, such as multilinguality \cite{conneau2019unsupervised, garcia2020harnessing, garcia2020multilingual,lin2021few}, back-translation \cite{lample2017unsupervised}, and most recently through large-scale models with parameters in the hundreds of billions \cite{chowdhery2022palm,vilar2022prompting,han2021unsupervised}. Despite such results, these systems are rarely compared to state-of-the-art supervised models which also leverage their own bag of tricks to improve performance.\footnote{For example, \citet{tran-etal-2021-facebook} applies model ensembling, reranking strategies, language-specific post-processing and fine-tuning on in-domain data for all language pairs, as well as iterative back-translation for the Icelandic language pairs.} Moreover, these models trained on large-scale monolingual data also suffer from similar issues as in models trained on large-scale parallel data: variance in the quality of monolingual data and generation being biased towards the over-represented registers in the monolingual data.

It thus remains to explore what lies between these two research streams: \emph{few-shot learning} \cite{brown2020language}. Recently, large language models have been shown capable of performing arbitrary tasks by exposing a few demonstrations of the task at inference time. The dependence of these models on demonstrations to perform the task allows us to overcome the aforementioned issues by carefully selecting a small set of high-quality translation pairs in the language register of interest as demonstrations.

In this work, we demonstrate that the few-shot translation paradigm allows us to build high-fidelity translation models at a smaller scale (8B parameters) than traditional large language models ($>100$B parameters) without the need for back-translation or large-scale parallel text mining. We evaluate these models on the WMT'21 English\minus{}German and English\minus{}Chinese news translation task and show that they outperform commercial baselines, and show performance competitive with WMT'21 submissions, which themselves rely on many of the aforementioned techniques. We then verify that our approach works in low-resource scenarios by performing a similar study on the WMT'21 English\minus{}Icelandic language pair, where the amount of Icelandic monolingual data is two orders of magnitude smaller than Chinese or German. Furthermore, we show that constraining the demonstrations to be in a desired language register generally results in the output translation being part of the register. We show that this results in quantitative gains in translation benchmarks which account for language registers: the region-aware benchmark FRMT \cite{riley2022frmt} and the IWSLT'22 Special Task on Formality Control for Spoken Language Translation \cite{anastasopoulos-etal-2022-findings}.

\begin{table}
\centering
\small
\begin{tabular}{cccccccll}
\toprule
Language & Examples \\ \midrule 
English & 69,813,325 \\
German &  69,813,325  \\
Chinese & 33,172,846 \\
Icelandic & 250,416 \\ \bottomrule
\end{tabular}
  \caption{\textbf{Number of processed examples per language.} Here by \emph{example}, we mean the tokenized inputs that are being passed to the model.  We describe the process used to create these examples in Section \ref{subsec:processing}. }
  \label{tab:data_comp}
\end{table}

\section{Experiments on high-resource languages \& results} \label{sec:main}
In this section, we outline the datasets, models, and initial set of experiments we perform in this work. We first discuss the composition of our monolingual data, as well as the evaluation datasets considered. Next, we describe the exact architecture we use for these studies, and how we train our models in terms of the particular objective being used and batching choice. Finally, we evaluate our few-shot translation models and compare their performance against a suite of state-of-the-art models. 

All the experiments in this work were conducted using JAX \cite{bradbury2021jax}, using the T5X framework \cite{roberts2022scaling} and FLAX \cite{flax2020github}. 
\subsection{Datasets}

\paragraph{Monolingual datasets} Our training data consists of a collection of language-specific corpora. For English, we use a similar mix of filtered web pages, Wikipedia, and books as done in \citet{chowdhery2022palm}. For every other language, we restricted ourselves to only high-quality webpages, using similar filters as the English data. The amount of data obtained by this approach varies by language. We list the final number of examples after processing in Table \ref{tab:data_comp}. We provide an explicit description of the processing in Section \ref{subsec:processing}. Collecting data in this process leads to much more English data than reported in our table. For simplicity, we artificially restricted the amount of data to be as much as the next highest-resource language, German.

\paragraph{Evaluation datasets} Most work in the unsupervised translation literature tends to focus on older WMT datasets e.g. WMT'14 English\minus{}French. This is problematic for us for a number of reasons: 1) the quality of WMT submissions have increased dramatically in recent years; 2) the quality of test sets have improved over the years; 3) because our pretraining data is derived from recent web crawls, there is a possibility of train / test overlap with previous years. For these reasons, we follow the recommendations outlined in \citet{vilar2022prompting} and use only recent test sets, coming from the WMT'21 news translation task. We primarily focus on English\minus{}German and English\minus{}Chinese language pairs, as these are typically high-resource language pairs, and thus we believe should have strong WMT submissions. 

\paragraph{Train-test overlap} To account for potential train-test overlap, we follow the strategy used in previous work \cite{chowdhery2022palm,vilar2022prompting} to measure target-side test overlap based on $n$-gram matching. We use 15-grams, with the understanding that test sequences shorter than 15 tokens will count as a match if they are found as a substring in the training data. We report the degree of overlap in Table \ref{tab:overlap} and note that we do not see much overlap with the newer test sets. 

\begin{table}
\centering
\small
\begin{tabular}{cccccccll}
\toprule
Language Pair & Forward & Backward \\ \midrule 
English-German &0.9\% & 1.5\% \\
English-Chinese &3.0\% & 1.5\% \\
English-Icelandic &0.3\% & 1.7\% \\ \bottomrule
\end{tabular}
  \caption{\textbf{Percent overlap between the references of each language pair and and the monolingual data.} We follow the same 15-gram protocol as \citet{chowdhery2022palm}. }
  \label{tab:overlap}
\end{table}

\subsection{Architecture and Training Procedure} \label{subsec:processing}

\paragraph{Architecture} We use a Transformer \cite{vaswani2017attention} decoder-only architecture, using the same architecture modifications as in \citet{chowdhery2022palm}. We use the exact hyperparameter configurations as their 8 billion parameter model for our main experiments. In particular, we use 32 Transformer layers, with 16 heads, a hidden dimension of 4096, and multi-query attention. The feed-forward size is 16384 and the attention head size is 256.

\paragraph{Vocabulary} We use a Sentencepiece \cite{kudo2018sentencepiece} model that was built in a similar fashion to the one used in \citet{chowdhery2022palm}, with the notable difference that we use 128,000 Sentencepieces instead 256,000. In particular, we use the same Sentencepiece model for all the models trained in this paper.

\paragraph{Data pre-processing and training objective} In this work, we use a variant of the UL2 objective \cite{tay2022unifying} that has been specialized for decoder-only models. The original UL2 objective relies on a mixture of 6 difference instances of the span corruption objective \cite{raffel2020exploring} using different hyperparameter configurations, coupled with a prefix language modeling \cite{raffel2020exploring}. The span corruption instances are completely determined by two hyperparameters: \emph{noise density}, which controls how much of the input is corrupted and \emph{mean noise span length}, which controls the average number of tokens corrupted per span. In this work, we use 2 (instead of 6) separate span corruption instances with (noise density, mean noise span length) given by (0.15, 3) and (0.5, 32) respectively. In addition to these two objectives and the prefix language modeling objective, we also include a standard causal language modeling objective. We mix these objectives randomly, sampling prefix language modeling 20\% of the time, causal language modeling 60\% of the time, and the remaining span corruption instances 20\% of the time.\footnote{The exact hyperparameter choices for this preprocessor were chosen following recommendations from the authors of \citet{tay2022unifying}. Early experiments showed that it performed better than just using causal language modeling and we did not explore the choice of hyperparameters any further.}

\begin{table*}
\centering
\small
\begin{tabular}{llccccccccccc}
\toprule
Rank & \emph{en} $\rightarrow$ \emph{zh} & \emph{zh} $\rightarrow$ \emph{en} & \emph{en} $\rightarrow$ \emph{de} & \emph{de} $\rightarrow$ \emph{en} \\ \midrule 
1st & \citet{zeng-etal-2021-wechat} & \citet{wang-etal-2021-tencent}  & \citet{tran-etal-2021-facebook} & \citet{tran-etal-2021-facebook} \\
2nd & \citet{tran-etal-2021-facebook}  & \citet{zhou-etal-2021-niutrans}  & \citet{qian2021volctrans} & Online\minus{}W   \\
3rd &\citet{zhou-etal-2021-niutrans} & \citet{li-etal-2021-miss-wmt21} & Online\minus{}W & \citet{subramanian-etal-2021-nvidia} \\ \bottomrule
\end{tabular}
  \caption{\textbf{WMT baselines for the high-resource language pairs we consider in this paper.} Note that although we cite the original paper, we recompute the metrics from the their textual outputs, rather than rely on the numbers provided in the paper to ensure fair comparison.}
  \label{tab:baselines}
\end{table*}

\paragraph{Trilingual models} We will be primarily focused on models exclusively supporting two languages at a time. However, to access the potential value of multilinguality in a more controlled setting, we will also consider \emph{trilingual models}, supporting three languages at a time. To establish a fair comparison between our trilingual and bilingual models, we need to enforce some constraints which should match practical use-cases. We enforce the constraint that the trilingual models must see the same amount of data per-language as the bilingual models i.e. the only difference in training data between the trilingual and bilingual models will be the additional data from the third language. While this choice results in more compute spent on training the trilingual models, it does not see any more data in any two languages compared to the analogous bilingual counterpart. 

\paragraph{Training distribution} For each set of languages considered, we perform one epoch over the combined corpus. This results in longer training for the English\minus{}German model than the English\minus{}Chinese model, due to the presence of more German data than Chinese data. We will assess the potential value of multi-epoch training in Section \ref{subsec:lr}.

\paragraph{Training hyperparameters} We use a maximum sequence length of 2048, with a batch size of 1024. We use the Adafactor optimizer \cite{shazeer2018adafactor}, without using the factorizing option. We use a cosine learning rate decay schedule \cite{hoffmann2022training}, starting at 0.01 and ending at 0.001 at the end of training. This results in 98000 steps for the English-Chinese model, 135000 steps for the English-German model, and 166000 steps for the trilingual model.

\paragraph{Evaluation} Our initial attempts at evaluating our models using BLEU \cite{papineni2002bleu} heavily underestimated the performance of our models. \citet{vilar2022prompting} has shown that translations generated from few-shot translation through large language models produce qualitatively different translations, with different kinds of errors from the ones generated by traditional machine translations models. We believe such errors are overly punished by the brittleness of n-gram metrics such as BLEU. Recent work has shown that such n-gram metrics are sub-optimal for evaluating high-quality translations \cite{kocmi-etal-2021-ship, freitag2021experts}.  For this reason, we use the learnt metric BLEURT \cite{sellam2020bleurt} as our main metric to assess quality. We follow the recommendations from the publicly-available Github page and use the BLEURT-20 checkpoint.\footnote{https://github.com/google-research/bleurt\#checkpoints} For completeness and full transparency, we report BLEU scores in the appendix, in Section \ref{sec:wmt_bleu}. 

\paragraph{Few-shot generation} To condition the model to perform translation, we use the relevant development set for each language pair as a pool of demonstrations. For each source example \texttt{text} in a given test set, we randomly sample 5 sentence pairs $(x_1, y_1), ... ,(x_5, y_5)$ from the development set and we insert these objects into the following template, where \texttt{source\_language} and \texttt{target\_language} are the names of source and target languages for the given language pair:
\begin{align*}
\{\texttt{source\_language}\} &: \{x_1\} \\
\{\texttt{target\_language}\} &:\{ y_1\} \\
\cdots\\
\{\texttt{source\_language}\} &: \{x_5\} \\
\{\texttt{target\_language}\} &:\{ y_5\} \\ \\
\{\texttt{source\_language}\} &: \{\texttt{text}\} \\
\{\texttt{target\_language}\} &: 
\end{align*}
To generate the predictions, we use minimum Bayes risk (MBR) decoding \cite{DBLP:journals/corr/abs-2005-10283,freitag2022high}, using learnt metrics and 64 sampled predictions with vanilla ancestral sampling. We use BLEURT for the utility function in MBR.  We also include results with beam search, with beam size 4 and $\alpha$=0.6.

\subsection{Main results and discussion}

\begin{table}
\centering
\small
\begin{tabular}{llccccccccccc}
\toprule
Model & \multicolumn{2}{l}{\stackanchor{\emph{en} $\leftrightarrow$ \emph{zh}}{\emph{newstest21}}} & \multicolumn{2}{l}{\stackanchor{\emph{en} $\leftrightarrow$ \emph{de}}{\emph{newstest21}}} \\ \midrule 
\textbf{Supervised baselines} \\
WMT'21 1st Place & \textbf{70.0}  & 66.6  & \textbf{76.9} & \textbf{76.9}  \\
WMT'21 2nd Place & 69.7  & 66.3  & 76.3 & 76.7   \\
WMT'21 3rd Place & 69.7 & 65.8  & 76.0 & 76.4 \\
Google Translate & 69.5  & 65.0  & 76.4 & 75.7 \\ \midrule
\textbf{Few-shot translation models} \\
PaLM & 67.7 & 64.1 & \underline{75.9} & 74.8 \\
\emph{Bilingual LMs (Beam)} & 62.6 & 67.0  &  74.9 & 74.1 \\ 
\emph{Bilingual LMs (MBR)} & 68.4 & 67.8  &  75.5 & 76.5 \\ 
 \emph{Trilingual LM (Beam)} & 65.3 & 65.3 & 74.5 & 74.4  \\ 
  \emph{Trilingual LM (MBR)} & \underline{68.9} & \underline{\textbf{68.3}} & 75.5 & \underline{76.8}  \\ \bottomrule
\end{tabular}
  \caption{\textbf{BLEURT scores from various models, both supervised and few-shot on some WMT \emph{newstest21} sets.} We italicized the name of our baselines, and bolded the best performing results. We also underline the best performing few-shot results. We use the suffix \emph{Beam} when using beam search, and \emph{MBR} when using MBR decoding.}
  \label{tab:hr_perf}
\end{table}
\paragraph{Baselines}  We consider the top three performing systems from WMT'21 for each language pair for comparisons. We provide a reference for each such system considered in Table~\ref{tab:baselines}. One potential pitfall of this approach is that WMT submissions have been hyper-specialized for this particular domain and evaluation procedure, and thus may overestimate the performance of general-purpose translation systems built to handle all domains. Since we are not focusing on making our translation models WMT-specific, we also include strong baselines of more general-purpose systems: \emph{PaLM} \cite{chowdhery2022palm}, a large multilingual (albeit English-centric) language model; \emph{Google Translate}, as an example of a commercial system which was not fine-tuned for WMT or any other such competition.\footnote{We re-used the translations from \cite{vilar2022prompting} when available. For the Icelandic pair, we obtained the predictions using the public API at translate.google.com in December 2022.} 

We list the performance of our models (labelled as \emph{Bilingual LMs} and \emph{Trilingual LM}) as well as the baselines on our dataset in Table \ref{tab:hr_perf}. To ensure fair comparison, we computed the BLEURT score directly from the text predictions of the baselines, as we did with our models. For PaLM, we used the predictions with the best performing metrics from \citet{vilar2022prompting}.\footnote{The PaLM numbers in Table \ref{tab:hr_perf} sometimes use different demonstrations from the ones we used. All our results draw demonstrations from the same development sets as in \citet{vilar2022prompting}, which they refer to as \emph{WMT-dev}. We only report PaLM numbers from \citet{vilar2022prompting} using different demonstrations if they are better than the ones obtained using WMT-dev as the source of demonstrations.} We first note that with the exception of German-English, both our language models outperform PaLM in this task when using MBR, despite having less than 2\% the number of parameters as PaLM. We believe this is in great part due to the larger number of monolingual data for the non-English languages seen in our models during training. This is especially highlighted in the out of English translation pairs (English\minus{}XX pairs.)

We also remark that most WMT baselines outperform the commercial system. As we noted earlier, this is most likely due to the fact that these baselines have been specialized for this WMT competition, while the commercial systems have to handle a more broader range of domains. When comparing the commercial system to our few-shot translation models, we note that our models excel in the English\minus{}XX directions despite also not being specialized for WMT. Moreover, our models are also able to outperform one of the strongest WMT baselines: for example, the trilingual model outperforms the best performing English\minus{}Chinese model from WMT'21, while being 0.1 BLEURT away from the best English\minus{}German system. The XX\minus{}English direction tells a different story: in this setting, we see our few-shot translations underperform against all our supervised baselines. Finally, we note that MBR consistently provides improvements in BLEURT. As such, we default to MBR for the remaining experiments.


\subsection{Performance on a low-resource language} \label{subsec:lr}

Finally, we consider whether this approach also yields high-quality translation models for low-resource languages. Building translation systems for such language pairs typically require leveraging data beyond the parallel data available for the language pair, whether in the form of monolingual data or parallel data for other language pairs. It is conceivable that the approach considered in this paper might not be as successful for languages with smaller amounts of monolingual data, as the value of parallel data could be magnified in these resource-constrained settings \cite{kirstain2021few}. 

We could attempt to study this problem by artificially constraining the amount of monolingual data for German or Chinese. However, previous works \cite{kim2020and,marchisio2020does} have shown that datasets for authentic low-resource languages offer additional difficulties that do not arise in datasets for high-resource languages, such as lower quality of data available, limited domain coverage, little language similarity with English, rich morphology, etc. To avoid this pitfall, we consider \emph{Icelandic} as our low-resource language of study.\footnote{Icelandic exhibits many of the properties typically associated with low-resource languages. For example, \citet{joshi-etal-2020-state} places Icelandic in the same category as Hausa, which WMT'21 considers to be low-resource.} Icelandic is also convenient for us since it is one of the languages available in WMT'21, and thus we have strong baselines available in the form of WMT submissions.  As can be seen from Table \ref{tab:data_comp}, Icelandic has two orders of magnitude less data than German or Chinese and hence poses a far more difficult challenge for our models.

\begin{table}
\centering
\small
\begin{tabular}{llcccccc}
\toprule
Rank & \emph{en} $\rightarrow$ \emph{is} & \emph{is} $\rightarrow$ \emph{en} \\ \midrule 
1st & \citet{tran-etal-2021-facebook}  & \citet{tran-etal-2021-facebook} \\
2nd & \citet{zhou-etal-2021-niutrans}  & Online\minus{}B     \\
3rd & \texttt{MANIFOLD} & \citet{zhou-etal-2021-niutrans}  \\ \bottomrule
\end{tabular}
  \caption{\textbf{WMT baselines for the Icelandic language pairs.} Note that although we cite the original paper, we compute the metrics from the their textual outputs, since many of these works do not report BLEURT numbers.}
  \label{tab:lr_baselines}
\end{table}

\paragraph{Warm-starting} Training models with low-resource languages can be a difficult endeavor due to issues regarding overfitting and imbalanced datasets. Previous work on transfer learning in language modeling shows that continued training on a new distribution yields similar results to training from scratch, albeit at a much faster rate \cite{hernandez2021scaling}. In this work, we follow this approach and fine-tune our English\minus{}German model. This allows us to leverage all the English knowledge our model has already accumulated during its original training run, which is critical due to the English-centric nature of our evaluation. 

\paragraph{Multi-epoch training} Previous literature on language modeling has found some success with repeated epochs over the training data \cite{taylor2022galactica}. We explore this question by training on a equal mixture of English and Icelandic until we go 8 epochs over the Icelandic dataset and evaluate the current checkpoint at the end of every epoch. We evaluate both the English\minus{}Icelandic and Icelandic\minus{}English directions on the Flores \cite{goyal2022flores} \emph{devtest} sets for Icelandic, average the two BLEURT scores, then take the best-performing checkpoint and evaluate it on the WMT \emph{newstest21} dataset. We select the optimal checkpoint occuring after 6 epochs over the Icelandic dataset. 

\paragraph{Training hyperparameters} We re-use all the same hyperparameters as when training the high-resource language models, with the exception of the learning rate: we set it to constant at 0.001, following recommendations from previous works \cite{xue2020mt5}.

\paragraph{Comparison with baselines} We list our WMT baselines in Table \ref{tab:lr_baselines} and show the results on both directions in Table \ref{tab:lr_perf}. We see that for both directions, the results are competitive with the WMT baselines, and in fact surpass the commercial baseline for the English-Icelandic direction. Moreover, we see that PaLM fails to provide high-quality translations as measured by BLEURT.

\begin{table}
\centering
\small
\begin{tabular}{llcccccc}
\toprule
Model & \multicolumn{2}{l}{\stackanchor{\emph{en} $\leftrightarrow$ \emph{is}}{\emph{newstest21}}} \\ \midrule 
\textbf{Supervised models} \\
WMT'21 1st Place & \textbf{77.2}  & \textbf{76.1} \\
WMT'21 2nd Place & 74.3  & 72.3     \\
WMT'21 3rd Place & 74.3 & 70.4  \\
Google Translate & 76.8  & 71.1  \\ \midrule
\textbf{Few-shot translation models} \\
PaLM & 61.7 & 59.5 & \\ 
\emph{Bilingual LMs (MBR)} & \underline{76.2} & \underline{72.0}  \\ \bottomrule
\end{tabular}
  \caption{\textbf{BLEURT scores from various models, both supervised and fewshot on the WMT \emph{newstest21} test sets involving Icelandic (\emph{is}).} To obtain these models, we fine-tuned the English-German models from Section \ref{sec:main} on a mixture on English and Icelandic, for a few epochs over the Icelandic monolingual data. }
  \label{tab:lr_perf}
\end{table}

\begin{figure}
\includegraphics[scale=0.140]{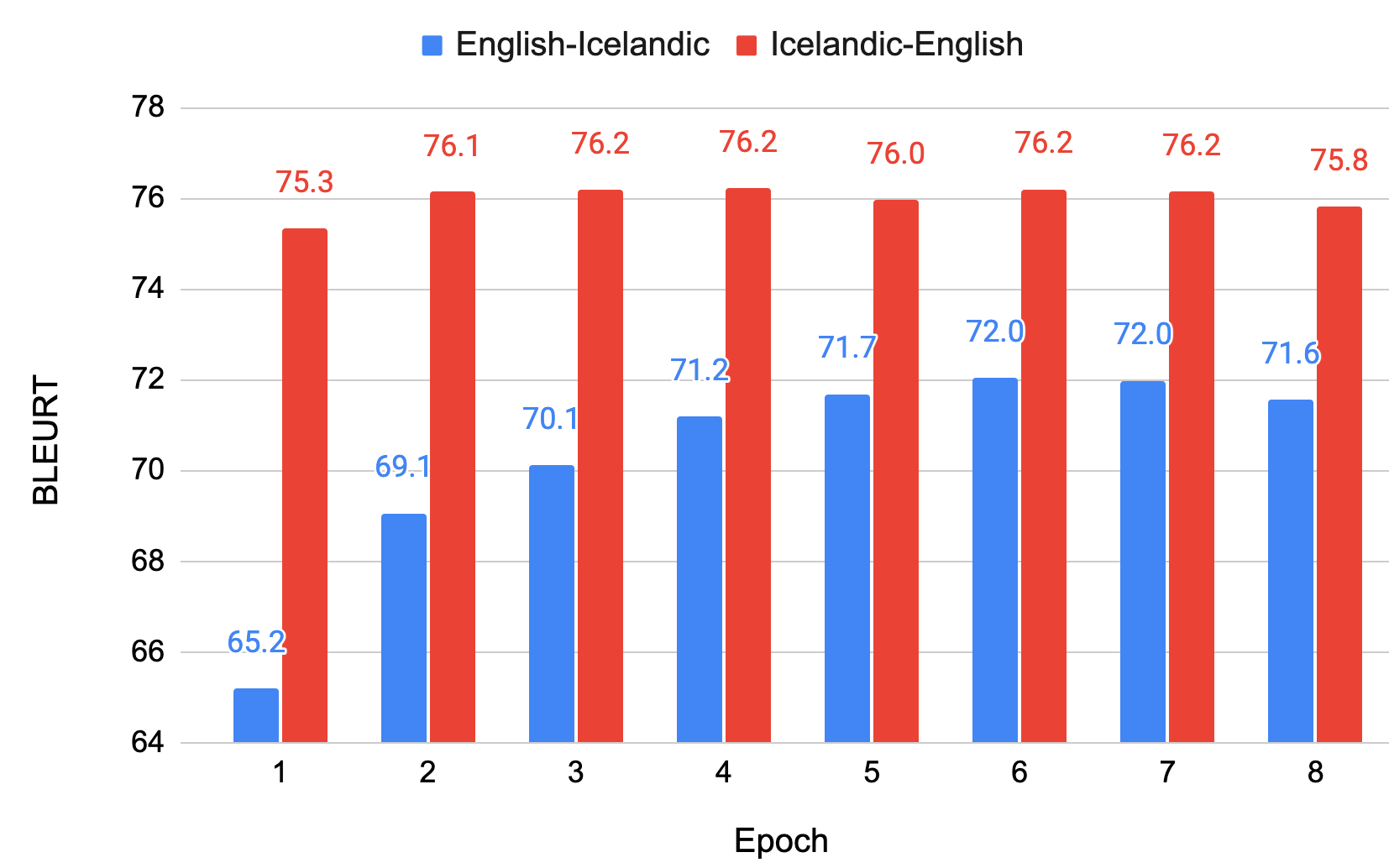}
\caption{\textbf{BLEURT scores on the WMT'21 English\minus{}Icelandic and Icelandic\minus{}English task after several epochs over the Icelandic monolingual data.} We remark that while the Icelandic-English direction converges fairly quickly (after two epoch), the English-Icelandic direction benefits from further training. }
\label{fig:multiepoch}
\end{figure}

\paragraph{Results of multi-epoch training} We present the results of evaluating at each epoch over the Icelandic datasets in Figure \ref{fig:multiepoch}. We see a few patterns emerge. First, for both directions, doing a single-pass is suboptimal. Second, the Icelandic-English direction doesn't benefit beyond two epochs over the Icelandic dataset, while the reverse direction benefits for up to 6 epochs. This suggests that models develop the ability to extract meaning from text much earlier than when they can reliably generate fluent text. Moreover, it also suggests that we may be underestimating the potential performance of our bilingual models if we had extended their training to include multiple epochs over the non-English datasets. Finally, we note that our models remain fairly competitive with the WMT'21 submissions and in fact outperform the commercial system in the English\minus{}Icelandic direction.

\section{Influence of few-shot demonstrations}

From previous experiments, it is clear that few-shot translation models can be competitive with state-of-the-art supervised models. In this section, we dissect which aspects of the few-shot demonstrations impact the results.  In this first subsection, we claim that one of the most influential predictors for the quality of our translations are the quality of the few-shot demonstrations. Next, we demonstrate that \emph{style} of the few-shot demonstration also influence the style of the translation, granting users the ability to output arbitrary styles at inference by only requiring a few demonstrations exhibiting that style and with no additional fine-tuning costs. We additionally show that using the incorrect style can result in a measurable regression in BLEURT, further highlighting the need for such systems. To investigate this, we look at two specific use-cases: controlling for Mandarin regional varieties, and formality in German. 
\subsection{Quality of the demonstrations strongly influences quality of generated translations} \label{subsec:quality}

Concurrent work \cite{agrawal2022context, vilar2022prompting} has shown that the quality of the few-shot demonstration impacts the quality of the output. We provide a similar analysis of this claim, by using a direct quantitative signal of the quality of the exemplar. 

\begin{figure}
\includegraphics[scale=0.2]{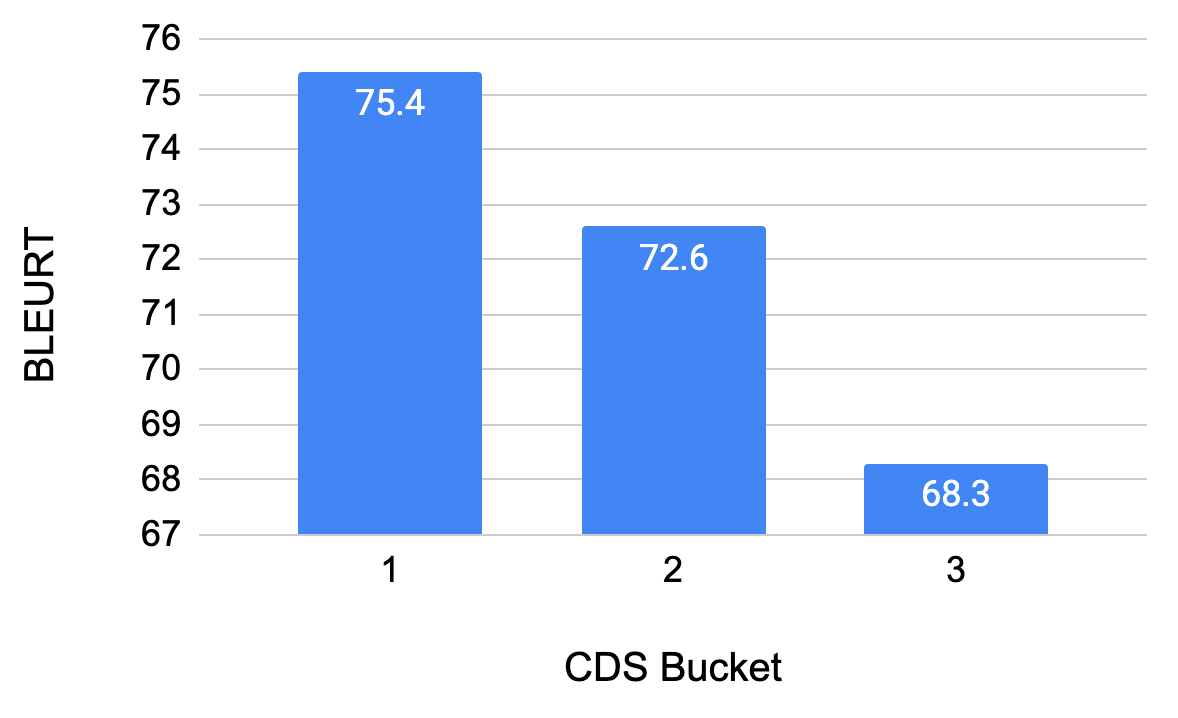}
\caption{\textbf{BLEURT scores on the WMT'21 English\minus{}German task, using demonstrations drawn of varying quality.} For each CDS bucket, we evaluate the model on the WMT'21 English\minus{}German task.}
\label{fig:cds}
\end{figure}

For this experiment, we diverge from our typical protocol of using development sets as our source of exemplars. Instead, we use an English\minus{}German translation dataset, prepared in \citet{bansal2022data}. This is a filtered version of the Paracrawl English\minus{}German dataset \cite{banon2020paracrawl}, where de-duplication has been applied, with both length and language identification filtering. This dataset also comes equipped with normalized Contrastive Data Selection (CDS) scores \cite{wang2018denoising}. These scores are at the example-level, and are calculated by computing the difference in cross-entropy scores between a translation model fine-tuned on a trusted dataset and one that was not. We should expect examples with higher CDS score to be increasingly different from the trusted data, and as such potentially of lower quality. The scores have been normalized to be between 0.0 and 1.0. We partition the examples into 3 buckets, where an example with a CDS score between 0.0 and 0.33 gets placed in the first bucket, an example with a CDS score between 0.33 and 0.66 gets placed in the second bucket, and the remaining examples go in the third bucket.  By viewing CDS scores as a proxy for quality, we can then examine the impact of quality of the demonstrations by constraining our demonstrations to come from a given bucket. We use the same decoding strategy as in our previous experiments, and report the BLEURT scores for each bucket in Figure \ref{fig:cds}. We note a pattern emerges: in general, as we use demonstrations with increasing CDS scores, the quality of the translations produced by our model keeps degrading.


\begin{table}
\centering
\small
\begin{tabular}{llccccccccccc}
\toprule
Model & \stackanchor{FRMT}{Score} & \stackanchor{Lexical}{Accuracy} \\ \midrule 
\textbf{FRMT Baselines} \\
PaLM 8B & 58.3 & 69.0 \\
PaLM 62B & 65.1 & 70.8  \\
PaLM 540B & 68.4 & 83.6 \\
Gold & \, \,\minus{}& 94.4 \\ \midrule 
\textbf{Ours} \\
\emph{Bilingual LM (mismatched)} & 66.0 & 13.8 \\
\emph{Bilingual LM (matched)} & \textbf{71.0} & \textbf{86.2} \\ \bottomrule
\end{tabular}
  \caption{\textbf{Performance of various models on the Mandarian regional varieties datasets from FRMT.} We use the term \emph{matched} and \emph{mismatched} for the cases when we use demonstrations from the same or different language variety as the references, respectively.  We bold the best results which do not include the reference. Note that Gold has no FRMT score, since this score is computed using the Gold predictions as references.}
  \label{tab:frmt}
\end{table}

\subsection{Controllability of output language variety through few-shot demonstrations} \label{subsec:regional varieties}

We now show that example quality is not the only attribute of the demonstrations which can heavily influence the quality of the outputs. We now claim that the \emph{style} of the demonstrations also influences the output in a quantitively-measurable way. This is exciting since it establishes the possibility of generating translations in a given style, such as formality-aware controlled translation \cite{niu2017study,niu2018multi,garcia2021towards} or targeting specific language varieties \cite{lakew2018neural,kumar2021machine}. Moreover, since the style is provided at inference, this allows us to condition the generation on arbitrary styles using only a very small set of high-quality translations exhibiting this attribute.

For this analysis, we use the FRMT dataset \cite{riley2022frmt}, which contains high-quality translations between English and two Mandarin regional varieties: Mainland and Taiwanese. We focus on two metrics: \emph{FRMT Score}, which consists of a geometric mean across regional varieties of the arithmetic means of the bucket-level BLEURT scores. As a proxy to measure whether the correct language variety is being used, we use a metric introduced in \citet{riley2022frmt} known as \emph{lexical accuracy}. To compute lexical accuracy, we follow the approach proposed by \citet{riley2022frmt} and use a special subset of the data (denoted in \citet{riley2022frmt} as the \emph{lexical bucket}) which has examples containing English terms that have different translations in the different regional varieties and check how often do the translations match the target language variety.

As in the previous experiments, we use the corresponding development sets to draw demonstrations for the translation task.\footnote{In addition to these demonstrations, we also use the names \emph{Mainland Mandarin} and \emph{Taiwanese Mandarin} for the respective regional varieties. } To highlight the importance of using demonstrations in the right language variety, we evaluate our models in two ways: \emph{matched}, in which we use the same language variety for the demonstrations as the reference; \emph{mismatched}, where we use different regional varieties for the demonstrations and the references. We compare against PaLM at various sizes, where the 540B model was the best-performing models in \citet{riley2022frmt}. We also include the lexical accuracy as human performance, which we refer as \emph{Gold}.

We report the results in Table \ref{tab:frmt}. We first note that even in the mismatched settings, our models still outperform the PaLM-62B model in terms of FRMT score, confirming the fidelity of the translations. However, we see that the lexical accuracy for the mismatched setting is very low, suggesting that the model is outputting valid translations but not in the target language variety. Once we use demonstrations showcasing the appropriate language variety, both the FRMT score and lexical accuracy drastically improve, establishing a new state-of-the-art, despite being two orders of magnitude smaller than the previous state-of-the-art.

\subsection{Controllability of formality through few-shot demonstrations} \label{subsec:formality}

Finally, we consider the task of generating translations satisfying a given formality level. While the formality of text is in general a subjective matter in English, other languages have built-in rules for expressing formality. For example, many languages exhibit a form of the T\minus{}V distinction \cite{brown-gilman-1960}, where the second person pronoun has two surface forms, an informal and a formal one. These pronouns usually also induce different verb conjugations, so even if they are not explicitly present in the sentence, one can still detect whether the intended style is informal or formal. 


In particular, we focus on the English\minus{}German language pair of the IWSLT'22 Special Task on Formality Control for Spoken Language Translation.\footnote{https://iwslt.org/2022/formality} We evaluate our models using both BLEURT and accuracy on whether the correct formality was used. To eliminate any discrepancies between our evaluation and the ones done by the organizers, we use the same evaluation scripts used by the official task for the formality accuracy.\footnote{https://github.com/amazon-science/contrastive-controlled-mt/blob/main/IWSLT2022/scorer.py} As in the language variety experiments, we consider both the \emph{matched} and \emph{mismatched} setting, where the demonstrations exhibit the same (respectively, different) formality level as the target set. We draw our demonstrations from the \emph{topical chat} split of the train set.\footnote{The choice of using the \emph{topical chat} split over \emph{telophony} split of the training set was arbitrary.}

\paragraph{Baselines} We consider Google Translate, which has no built-in feature for controlling formality. We also consider the \texttt{gold finetuned mBART-large} UMD submission \cite{rippeth-etal-2022-controlling} for the task.\footnote{The explicit text predictions can be found here: \text{https://github.com/amazon-science/contrastive-controlled-mt/}
\textnormal{tree/main/IWSLT2022/submissions/formality-control\_UMD/EN-DE/unconstrained/formality/blind-test}} We choose this submission from UMD as it has the highest translation quality compared with UMD's other submissions.\footnote{There are other submissions with better formality control than the submission we selected. However, we believe translation quality should be prioritized over control of stylistic features such as formality and hence we selected this submission over the alternatives.}

We report the BLEURT scores and formality accuracy in Table \ref{tab:iwslt2022}. We first note that even though we took UMD's submission with the highest translation quality, it still attains lower quality than all the other models, including our mismatched models. However, UMD's submission exhibits higher formality accuracy compared to the commercial baseline, suggesting that there might be a trade-off in controlling formality versus quality. Our few-shot translation models achieve the best of both worlds: they exhibit high-quality translations, outperforming all available models, as well as achieving excellent formality control, obtaining higher accuracy than the UMD submission on the informal test set, despite only seeing a five examples at inference. Moreover, we do see increased BLEURT scores using the matched exemplars over the mismatched ones, suggesting that BLEURT is sensitive to formality. 

\begin{table}
\centering
\small
\begin{tabular}{llccccccccccc}
\toprule
Model &  \multicolumn{2}{l}{\stackanchor{BLEURT}{Formal\minus{}Informal}}& \multicolumn{2}{l}{\stackanchor{Accuracy}{Formal\minus{}Informal}} \\ \midrule
\textbf{External Baselines} \\
Google Translate & 72.2 & 72.5 & 62.7\% & 37.3\% \\ 
UMD  & 69.4 & 69.3 & \textbf{93.6\%} & 77.4\% \\ \midrule
\textbf{Ours} \\
\emph{Mismatched} & 72.1 & 72.2 & 15.1\% & 15.5\% \\
\emph{Matched} & \textbf{73.5} & \textbf{74.7} & 84.9\% & \textbf{85.5\%} \\ \bottomrule
\end{tabular}
  \caption{\textbf{Performance of various models on the IWSLT2022 Formality Control for Spoken Language Translation.} We use the term \emph{matched} (respectively, \emph{mismatched}) for the cases when we use demonstrations from the same (respectively, different) formality as the references.  We bold the best results and italicize the name of our models.}
  \label{tab:iwslt2022}
\end{table}

\section{Limitations and Future Work}

Despite our strong results, we note that there are a few limitations of the approach described in this work which we would like to address in future work. 

\paragraph{Model size} While we were able to outperform large language models like PaLM while using only a fraction of the number of parameters, our resulting models are still an order of magnitude larger than traditional machine translation systems, which usually have hundreds of millions of parameters. Preliminary experiments at the 1 billion parameter scale yielded unsatisfactory results, which is why we restricted our experiments to 8 billion parameter scale. Such size limitations may inhibit adoption of these techniques and future work should aim at lowering these size requirements.

\paragraph{Multilinguality} Our initial trilingual experiments showed that our models benefited from multilinguality. However, previous work in the multilingual literature has shown that adding more languages eventually leads to a decrease in per-language performance, a phenomenon known as the \emph{curse of multilinguality} \cite{conneau2019unsupervised}. Future work should explore how many languages can such systems support until we encounter performance degradation. 

\paragraph{Incorporating larger amounts of high-quality parallel data} In Section \ref{subsec:quality}, we demonstrated that one of the critical properties of parallel data that we needed was \emph{quality}, in contrast to the current popular approaches used in supervised models that mainly rely on \emph{quantity}. However, our approach only utilizes five examples at inference, even though one could potentially get several hundreds of high-quality examples. If we were to obtain such a dataset and fine-tune our models using traditional approaches, we might obtain higher-quality translation models, but then we might also lose the exciting controllability that we get through the few-shot translation paradigm as shown in Sections \ref{subsec:regional varieties} and \ref{subsec:formality}. Future work should look into developing approaches that can leverage larger sets of high-quality data while still retaining the flexibility of the few-shot translation paradigm.

\paragraph{Decoding algorithm} To get our best results, we relied on MBR decoding rather than beam search. We believe this to be a fair comparison with the WMT submissions, since they also rely on various reranking strategies. On the other hand, \citet{vilar2022prompting} did not rely on MBR decoding for their results, and we should expect PaLM's translation quality to improve when using MBR. Moreover, using these kind of decoding algorithms incurs a larger cost for serving than beam search, further inhibiting adoption. Future work should try to recover similar performance without the need for expensive decoding algorithms.

\section{Conclusion}

In this work, we investigated the potential value of few-shot translation models by examining their performance against the strongest supervised baselines we could find on traditionally high-resource language pairs. We show that these systems can be competitive with supervised models under the assumption that we have a small set of high-quality demonstrations. We analyze how the quality of these demonstrations heavily influences the quality of the translations generated by our models. Finally, we demonstrate that this paradigm also gives a way of controlling the style of the translations generated by these models. 

\bibliography{references}
\bibliographystyle{icml2022}

\newpage
\appendix
\onecolumn
\section{BLEU scores for the WMT'21 Language Pairs} \label{sec:wmt_bleu}

\begin{table}
\centering
\small
\begin{tabular}{llcccccccccc}
\toprule
Models & \multicolumn{2}{l}{\stackanchor{\emph{zh} $\leftrightarrow$ \emph{en}}{\emph{newstest21}}} & \multicolumn{2}{l}{\stackanchor{\emph{de} $\leftrightarrow$ \emph{en}}{\emph{newstest21}}} & \multicolumn{2}{l}{\stackanchor{\emph{is} $\leftrightarrow$ \emph{en}}{\emph{newstest21}}} \\ \midrule
\textbf{Supervised models} \\

WMT'21 1st Place & 33.4 & 36.9 & 41.9 & 42 & 41.7 & 33.3 \\ 
WMT'21 2nd Place & 31.9 & 35.9 & 39.7 & 43.2 & 40 & 30.6 \\ 
WMT'21 3rd Place & 32.6 & 35.8 & 40 & 41.3 & 39.2 & 28.6 \\ 
Google Translate & 32.2 & 36.2 & 40.9 & 39.8 & 41.5 & 28.7 \\ \midrule
\textbf{Few-shot translation models} \\
PaLM & 25.8 & 29.6 & 38.8 & 32.9 & 19.1 & 16.87 \\ 
\emph{Bilingual LM (MBR)} & 21.6 & 25.6 & 37.1 & 29.4 & 33.52 & 17.7 \\ 
\emph{Bilingual LM (Beam)} & 20.4 & 29.2 & 35.5 & 32.8 & 36.2 & 19.2 \\ 
\emph{Trilingual LM (MBR)} & 22.2 & 26.8 & 36.1 & 28.5 & \minus{} & \minus{} \\ 
\emph{Trilingual LM (Beam)} & 20.45 & 25.5 & 36.2 & 31.8 & \minus{}  & \minus{}  \\ \bottomrule
\end{tabular}
  \caption{\textbf{BLEU scores from various models, both supervised and few-shot on some WMT \emph{newstest21} sets.} We italicized the name of our baselines. We use the suffix \emph{Beam} when using beam search, and \emph{MBR} when using MBR decoding. Note that we have no scores for the trilingual models on the Icelandic pairs since we didn't train any such models.}
  \label{tab:wmt_bleu}
\end{table}

We list out the BLEU scores, computed using SacreBLEU.\footnote{SacreBLEU signature: nrefs:1|case:mixed|eff:no|tok:TOK|smooth:exp|version:2.1.0, where TOK is 13a or zh.
} We note that there is a big gap between the WMT models and our models. On one hand, \citet{freitag2022high} reported that BLEU drops sharply when using MBR. However, we also find this pattern even when using beam search, suggesting that few-shot translation models naturally produce translations which are qualitatively different from traditional supervised models. 

\section{Model architecture} \label{sec:model_arch}

We use 32 Transformer layers, with 16 heads, a hidden dimension of 4096, and multi-query attention. The feed-forward size is 16384 and the attention head size is 256.


\end{document}